\documentclass{article}

\usepackage{arxiv}

\usepackage[utf8]{inputenc} 
\usepackage[T1]{fontenc}    
\usepackage[hidelinks]{hyperref}      
\usepackage{url}            
\usepackage{booktabs}       
\usepackage{amsfonts}       
\usepackage{nicefrac}       
\usepackage{microtype}      
\usepackage{lipsum}		
\usepackage{graphicx}
\usepackage{natbib}
\usepackage{doi}
\usepackage{amsmath} 
\usepackage{multirow}

\usepackage{algorithmic}
\usepackage{algorithm}

\title{The Estimation of Continual Causal Effect for Dataset Shifting Streams \thanks{We thank Jianting Guo and Xuegang Kong for their support, collegiality and collaboration during the project.}}

\author{ {Baining Chen} \\
	Didi Chuxing\\
	\texttt{chenbaining\_i@didiglobal.com} \\
	\And
	{Yiming Zhang}\\
	Didi Chuxing\\
	\texttt{eddiezhang@didiglobal.com} \\
    \And
	{Yuqiao Han} \\
	Didi Chuxing\\
	\texttt{hanyuqiao@didiglobal.com} \\
    \And
    {Ruyue Zhang} \\
	Didi Chuxing\\
	\texttt{zhangruyue@didiglobal.com} \\
    \And
    {Ruihuan Du}\thanks{Corresponding author} \\
	Didi Chuxing\\
	\texttt{duruihuan@gmail.com} \\
    \And
    {Zhishuo Zhou} \\
	Didi Chuxing\\
	\texttt{zhouzhishuo@didiglobal.com} \\
    \And
    {Zhengdan Zhu} \\
	Didi Chuxing\\
	\texttt{zhuzhengdan@didiglobal.com} \\
    \And
    {Xun Liu} \\
	Didi Chuxing\\
	\texttt{wubusi@didiglobal.com} \\
    \And
    {Jiecheng Guo} \\
	Didi Chuxing\\
	\texttt{jasonguo@didiglobal.com} \\
}

\hypersetup{
pdftitle={ICE-PKD},
pdfauthor={Baining Chen},
}
\begin{document}
\maketitle

\begin{abstract}
Causal effect estimation has been widely used in marketing optimization. The framework of an uplift model followed by a constrained optimization algorithm is popular in practice. To enhance performance in the online environment, the framework needs to be improved to address the complexities caused by temporal dataset shift. This paper focuses on capturing the dataset shift from user behavior and domain distribution changing over time. We propose an Incremental Causal Effect with Proxy Knowledge Distillation (ICE-PKD) framework to tackle this challenge. The ICE-PKD framework includes two components: (i) a multi-treatment uplift network that eliminates confounding bias using counterfactual regression; (ii) an incremental training strategy that adapts to the temporal dataset shift by updating with the latest data and protects generalization via replay-based knowledge distillation. We also revisit the uplift modeling metrics and introduce a novel metric for more precise online evaluation in multiple treatment scenarios. Extensive experiments on both simulated and online datasets show that the proposed framework achieves better performance. The ICE-PKD framework has been deployed in the marketing system of Huaxiaozhu, a ride-hailing platform in China. 
\end{abstract}


\section{Introduction}
Causal effect estimation is critical in marketing optimization \citep{he2024rankability, sun2015causal, liu2023explicit} and is applicable in various other domains, including healthcare \citep{smit2023causal, sato2024uplift} and finance transactions \citep{kristjanpoller2025incorporating, kumar2023causal}. Previous works have made considerable progress in causal effect estimation using observational data, as randomized controlled trial (RCT) data is often costly and difficult to obtain.

To address the challenges posed by missing counterfactual outcomes and selection bias \citep{vemuri2015causal}, several approaches have been proposed, including re-weighting \citep{austin2011introduction, bottou2013counterfactual,zubizarreta2015stable,hassanpour2019counterfactual}, balanced representation \citep{bengio2013representation, shalit2017estimating} and disentangled representation \citep{hassanpour2019counterfactual, wu2022learning}.

The training framework must be updated when the distribution of online source data changes over time. However, above approaches do not adequately address this issue. As a result, the model's performance decreases very quickly over time \citep{chu2023continual}. In online marketing, data sources are influenced by several factors, including news, holidays and weather conditions. These temporal changes in data are referred to as dataset shift. In practice, two kinds of dataset shift are commonly considered, which are covariate shift \citep{nair2019covariate} and concept drift \citep{kull2014patterns}.

\textbf{Covariate shift} is the temporal changes of the covariates' distribution. 

\textbf{Concept drift} is the temporal changes in the relationship between the covariates and the responses (or outcomes) for each treatment group.

Taking Huaxiaozhu (a popular ride-hailing platform in China) as an example: the temporal changes in user profile distribution can be viewed as covariate shift, and the temporal changes in how user profiles influence the orders can be interpreted as concept drift.

Continual Causal Effect Representation Learning method (CERL) is proposed \citep{chu2023continual} to address the problem of continual lifelong causal effect estimation. The CERL focuses on covariate shift in binary treatment scenarios. However, our business needs to face the challenges of both covariate shift and concept drift in multiple treatment settings.

In this paper, we propose the Incremental Causal Effect with Proxy Knowledge Distillation (ICE-PKD), a framework designed to handle multiple treatments and effectively address the complexities associated with concept drift. ICE-PKD provides a more robust and adaptable approach for causal effect estimation in dynamic marketing settings. Specifically, ICE-PKD extends the Disentangled Representations for CounterFactual Regression (DR-CFR)  framework \citep{hassanpour2019learning} into multiple treatments and draws inspiration from class-incremental methods in computer vision \citep{van2022three, belouadah2021comprehensive}, such as functional regularization \citep{li2017learning} and relay-based techniques \citep{rebuffi2017icarl}. Additionally, we introduce a customized modification to the knowledge distillation process by incorporating insights from the alternative method of PTLoss \citep{zhang2024knowledge}, addressing concept drift more effectively. As a result, ICE-PKD is able to learn the continual causal effect for dataset shifting streams. The effectiveness of ICE-PKD has been validated via extensive experiments.

The main contributions of this paper are as follows: (1) an extended DR-CFR framework for handling multiple treatments; (2) a novel training strategy for continual causal effect estimation, considering both covariate shift and concept drift; (3) a new metric RAS-AUCC which evaluates the uplift model in multiple treatments.

\section{Related Work}
In this section, we present a brief overview of relevant works, including uplift modeling, evaluation metrics for uplift modeling, incremental learning, knowledge distillation with auxiliary loss terms and continual causal effect estimation.

\subsection{Uplift Modeling}
\label{section:uplift modeling}
Uplift modeling focuses on estimating the Individual Treatment Effect (ITE) and has undergone significant advancements over time. Existing methods can be broadly classified into three main types: meta-learner-based methods, tree-based methods and neural network-based methods.  

\paragraph{Meta-learner-based methods.} These methods use machine learning estimators as base learners to estimate treatment effects by comparing predicted responses. The S-learner \citep{lo2002true} trains a single model on the combined dataset of treated and control samples, incorporating the treatment indicator as an additive input feature. In contrast, T-learner \citep{kunzel2019metalearners} constructs two distinct models: one for the treated group and one for the control group.

\paragraph{Tree-based methods.} These methods leverage specialized tree or forest structures with treatment-effect-optimized splitting criteria. Bayesian Additive Regression Trees (BART) \citep{chipman2010bart} merge the interpretability of decision trees with a Bayesian framework, capturing complex relationships and uncertainty within the data. Causal Forest \citep{wager2018estimation} is specifically designed to estimate heterogeneous treatment effects by employing random forests within a causal inference framework.

\paragraph{Neural network-based methods.} These methods fully leverage the approximation capability and structural flexibility of neural networks. Balancing Neural Network (BNN) \citep{johansson2016learning} minimizes the discrepancy between the distribution of control and treatment samples using the Integral Probability Metric (IPM). Treatment-Agnostic Representation Network (TarNet) and Counterfactual Regression (CFR) \citep{shalit2017estimating} adopt a multi-task network architecture, learning separate networks based on a shared balanced representation across the treated and control groups. DragonNet \citep{shi2019adapting} is designed to predict propensity scores and conditional outcomes through an end-to-end procedure. Disentangled Representations for CounterFactual Regression (DR-CFR) \citep{hassanpour2019learning} employs representation learning to design a model architecture that explicitly identifies underlying factors in observational datasets. Deep Entire Space Cross
Networks for Individual Treatment Effect Estimation (DESCN) \citep{zhong2022descn} addresses treatment bias and sample imbalance by jointly learning treatment and response functions across the entire sample space and utilizing a cross network to model hidden treatment effects. Explicit Feature Interaction-aware Uplift Network (EFIN) \citep{liu2023explicit} enhances online marketing by explicitly modeling treatment features and their interactions to accurately estimate individual treatment effects. 

\subsection{Evaluation Metrics for Uplift Modeling}
A significant challenge in uplift modeling is the accurate evaluation of model performance, primarily due to the unobservable nature of counterfactual outcomes in online datasets. The Precision in Estimating Heterogeneous Effects (PEHE) \citep{hill2011bayesian} is commonly employed in semi-synthetic or simulated datasets. It quantifies the discrepancy between ground truth and predicted treatment effects at the individual level. However, it necessitates knowledge of both counterfactual and factual outcomes, making it unsuitable for direct evaluation on online data. The Average Treatment Effect Error (ATE Error) measures the average difference in outcomes between treated and control groups. The Area Under the Uplift Curve (AUUC) \citep{devriendt2020learning} assesses a model's ability to rank individuals based on predicted uplift. The QINI Coefficient \citep{devriendt2020learning} introduces a scaling correction to the AUUC calculation, ensuring a more accurate assessment when the treatment and control groups are misaligned. The Area Under Cost Curve (AUCC) \citep{du2019improve} is similar to AUUC but specifically evaluates performance in marketing contexts based on Return On Investment (ROI). Furthermore, the Area Under Cost Curve for Multiple Treatments (MT-AUCC) \citep{zhou2023direct} extends AUCC to scenarios involving multiple treatments. 

These existing metrics primarily focus on the performance of uplift modeling. However, our main concern is online revenue. Therefore, we propose the Area Under Cost Curve for Resource Allocation Simulation (RAS-AUCC), a metric that more accurately reflects the overall effectiveness of online strategies.

\subsection{Incremental Learning}
Incremental learning \citep{de2019continual,van2022three,belouadah2021comprehensive} focuses on enabling models to learn from non-stationary data streams, a concept widely applied in computer vision. A significant challenge in incremental learning is catastrophic forgetting, where previously acquired knowledge is overwritten as new information is introduced. To mitigate this issue, incremental learning methods can be categorized into three primary families \citep{de2019continual}: parameter isolation methods, regularization-based methods and replay methods. 

\paragraph{Parameter isolation methods.} These methods allocate specific model parameters to different tasks, thereby preventing interference between them. Context-dependent Gating (XdG) \citep{masse2018alleviating} employs separate gates to control which parts of the model are used for each task. Hard Attention to the Task (HAT) \citep{serra2018overcoming} incorporates task-specific embeddings to apply attention masking, facilitating selective knowledge sharing between tasks.  

\paragraph{Regularization-based methods.} These methods aim to retain previous knowledge while incrementally updating the model through regularization. Elastic Weight Consolidation (EWC) \citep{kirkpatrick2017overcoming} introduces a penalty to the loss function to restrict significant changes to important weights. Synaptic Intelligence (SI) \citep{zenke2017continual} is akin to EWC but focuses on estimating weight importance based on their contribution to task performance. Learning without Forgetting (LwF) \citep{li2017learning} utilizes knowledge distillation to preserve knowledge from previous tasks while accommodating new ones. 

\paragraph{Reply methods.} These methods involve storing and revisiting previous data or representations to help the model retain previous knowledge while learning new tasks. Averaged Gradient Episodic Memory (A-GEM) \citep{chaudhry2018efficient} employs a memory buffer to store and replay a subset of past data. Incremental Classifier and Representation Learning (iCaRL) \citep{rebuffi2017icarl} selects and retains samples closest to the representation mean of each class, minimizing the loss on new classes while performing distillation on the replayed samples.

\subsection{Knowledge Distillation Enhanced by Auxiliary Loss Techniques}
Knowledge Distillation (KD) \citep{gou2021knowledge} is a technique in which a student model is trained to replicate the output or intermediate representations of a teacher model. It enables the student to achieve improved performance while maintaining computational efficiency. However, \citet{zhang2024knowledge} highlights that this learning objective can be sub-optimal due to discrepancies between the teacher's output distribution and the ground truth distribution. To address this issue, PTLoss \citep{zhang2024knowledge} transforms the vanilla teacher model into a proxy teacher model by perturbing the leading-order terms in the Maclaurin series of the logarithmic term. PTLoss provides a generalized framework that incorporates various techniques, including label smoothing \citep{szegedy2016rethinking}, temperature scaling \citep{hinton2015distilling} and focal loss \citep{lin2017focal}, through the manipulation of perturbation coefficients. Both theoretical and experimental results demonstrate the effectiveness of PTLoss. Furthermore, \citet{zhang2024knowledge} proposes an alternative solution that offers valuable insights for enhancing the knowledge distillation process, making it more adaptable to the scenarios discussed in this paper. 

\subsection{Continue Causal Effect Estimation}
Continual causal effect estimation involves estimating the causal effects of treatments in dynamic settings, where data evolves over time. The challenges associated with this estimation include not only counterfactual inference in the presence of selection bias but also the need to adapt to changes in the dynamic environment. Continual Causal Effect Representation Learning (CERL) \citep{chu2023continual} is the first attempt to address these issues. It builds upon the CFR framework \citep{shalit2017estimating}, incorporating an elastic net regularization term and 
feature representation distillation. Extensive experiments demonstrate its effectiveness in handling covariate shift within observational data streams with binary treatment. However, real-world situations are often more complex. The scenarios involving multiple treatments and concept drift are common in marketing. CERL does not account for these complexities, and this paper aims to address these challenges.

\section{Problem Statement}
Let \( X \) denote the set of covariates (variables influencing the outcomes), \( Y \) denote the set of binary outcomes, \( T \) denote the set of treatments, which consists of  \(0,1, \cdots,|T| \), and \( K \) denote the set of time periods, which consists of \(0,1, \cdots,|K| \). In continual causal effects estimation, observational data are collected across different time periods. The $k$-th dataset comprises the data \( \{(x,y,t) \mid x \in X, y \in Y, t \in T \} \) collected during the $k$-th time period, which contains \( N_k \) units. Due to covariate shift and concept drift, the datasets \( \{ D_k \}_{k=0}^{|K|} \) exhibit different distributions with respect to \( X \) and varying relationships between \( X \) and \( Y \). 

Covariate shift can be expressed as:
\[
P(x \mid k) \neq P(x \mid k-1) \quad \text{for } k = 1, \cdots,|K|
\]
Concept drift can be expressed as:
\[
P(y \mid x, t, k) \neq P(y \mid x, t, k-1) \quad \text{for } k = 1, \cdots,|K|
\]
Let \( t_i \) denote the treatment assignment for unit \( i \). For multiple treatments, \( t_i=0 \) corresponds to the control group, while \( t_i=1,, \cdots,|T|\) corresponds to the treatment groups. The outcome for unit \( i \) is denoted by \( y_i^t \) when treatment \( t \) is applied to unit \( i \). In observational data, only one of the potential outcomes is observed, referred to as the factual outcome, while all the remaining unobserved potential outcomes are termed counterfactual outcomes. For each time period (except \( 0 \)), the responses \( \hat{y}_i \) of treatments \( t=0,1,\dots,|T| \) applied to the unit \( i \) are predicted to estimate the expected Individual Treatment Effect (ITE) \( \tau_i^t \) under the treatments \( t=1,2,\dots,|T| \). Specifically, under three standard assumptions \citep{rosenbaum1983central, zhong2022descn}, the estimate of \( \tau_i^t \) can be expressed as follows:

\begin{equation}
    \tau_i^t=\mathbb{E}(y_i^t-y_i^0|x_i)=\mathbb{E}(y_i^t|t,x_i)-\mathbb{E}(y_i^0|t,x_i)
\end{equation}

Here, \( \mathbb{E}[y^{t}|t,x] \) represents the expected predicted response \( \hat{y}_i^t \) of individual \( x_i \) under treatment \( t \), for \( t=0,1,\ldots,|T|\). In scenarios such as marketing campaigns, the outcome \( y_i^t \in \{0,1\} \) typically indicates whether the individual made a purchase, while the predicted response \( \hat{y}_i^t \in [0,1] \) represents the probability of purchase for that individual.

Our objective is to develop a novel continual causal effect inference framework that accurately estimates causal effects from incrementally available data \( D_k \) in the presence of covariate shift and concept drift.

\section{Methodology}
As discussed in Section \ref{section:uplift modeling}, most existing methods inadequately capture the dynamics prevalent in many industrial applications. To address this challenge, we propose the Incremental Causal Effect with Proxy Knowledge Distillation (ICE-PKD), an efficient framework for performing uplift modeling on incrementally available data. The architecture of ICE-PKD is illustrated in Figure \ref{alg:ice-pkd}. Specifically, DR-CFR serves as the base model of ICE-PKD, providing robust causal inference capabilities in multi-treatment observational data scenarios. The initial model selection ensures reliable generalization before incremental training, while the knowledge distillation process preserves this generalization during adaptation to new changes. To mitigate the influence of covariate shift, we use the replay mechanism to maintain the accuracy and stability of the knowledge distillation process by optimizing the distillation process on the replayed dataset. To mitigate the influence of concept drift, we transform the original teacher model into a proxy teacher to improve the accuracy and reliability of the knowledge distillation by mitigating the discrepancy between the teacher model's output distribution and ground truth distribution. 

\begin{figure}
    \centering
    \includegraphics[height=6cm]{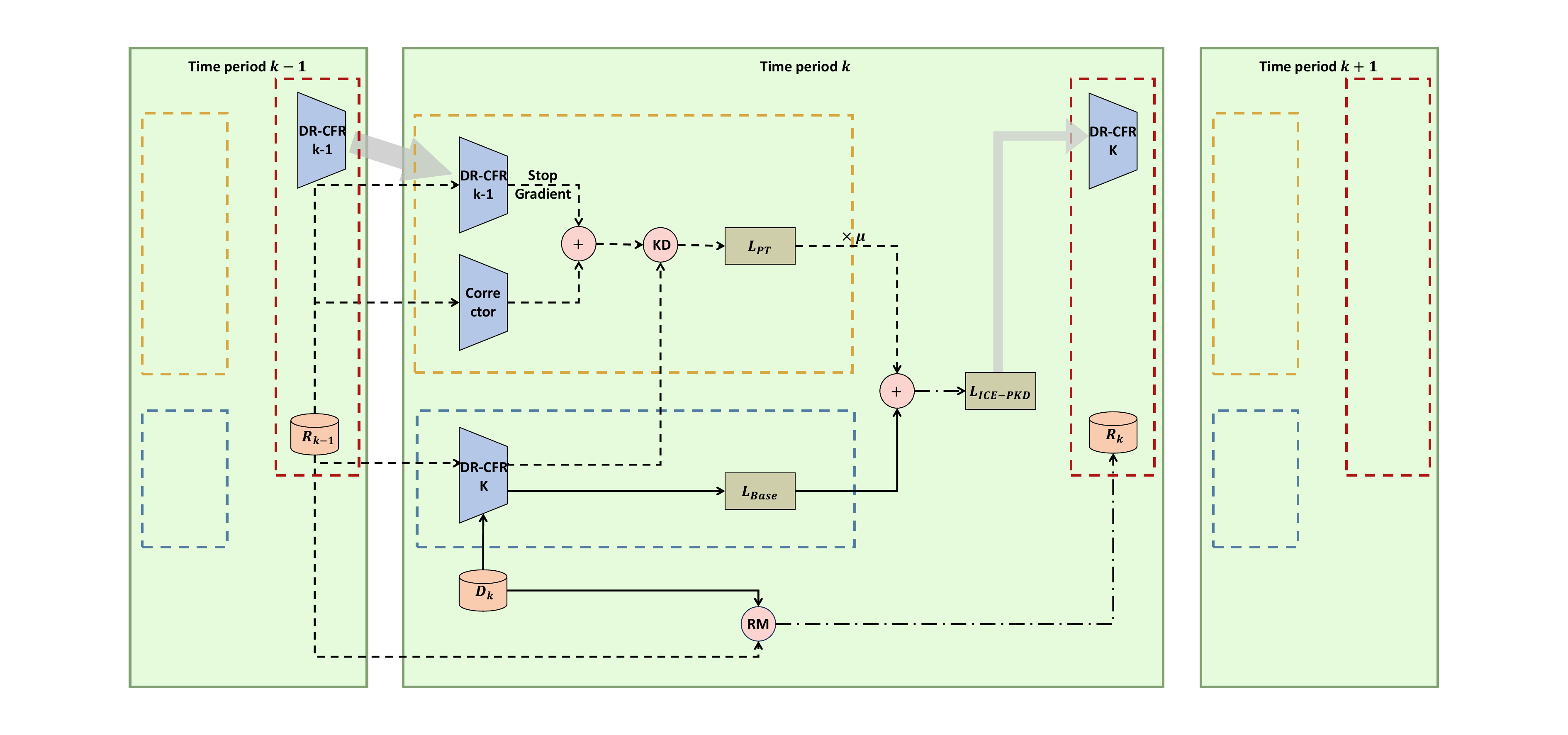} 
    \caption{Architecture of the ICE-PKD framework. The blue trapezoids denote the base model, orange bars represent the data, pink circles indicate the operators, yellow-gray rectangles correspond to the loss function. The black dashed lines illustrate the processing of thereplayed dataset, while the black solid lines represent the processing of the latest available dataset.}
    \label{fig:ice-pkd}
\end{figure}

\subsection{The Base Causal Effect learning Model}
In this section, we describe the base model for causal effect estimation in observational datasets. A key challenge in estimating causal effects from observational data is selection bias, where the distribution of data varies across treatment groups. This issue is obvious when there are few sample instances within a given domain for each treatment group, complicating precise inferences about counterfactual outcomes.

To address this challenge, we employ the DR-CFR framework \citep{hassanpour2019learning}. This framework disentangles covariates into underlying factors and balances the representation of confounders between control and treatment groups. This method is effective in enhancing the accuracy of causal effect estimation. In this paper, we extend it to accommodate multiple treatment scenarios while satisfying the monotonicity condition.

\paragraph{Learning Disentangled Representations.}
We assume that all covariates influencing the effect of the model are generated by three underlying factors $\{ \Gamma,\Delta,\Upsilon \}$ \citep{hassanpour2019learning}. The instrumental factor $\Gamma$ affects only the treatment assignment, the confounding factor $\Delta$ influences both the treatment and the outcome, and the adjustment factor $\Upsilon$ determines only the outcome. While the presence of instrumental and confounding factors introduces selection bias, confounding factors can still provide useful information for accurately estimating the response. To model these factors, we employ three representation learning networks \( \Gamma(x) \), \(\Delta(x) \) and \( \Upsilon(x) \), which learn the respective factors from data.

\paragraph{Re-Weighting Function.}
To address the selection bias in estimating Individual Treatment Effects, \citep{hassanpour2019learning} employs context-aware importance sampling, where the weights are derived exclusively from \(\Delta(x)\). In this work, we extend this approach to accommodate the multi-treatment scenario:
\begin{equation}
    w(t_{i},\Delta(x_{i}))=1+\sum_{j\ne i} \frac{p(t_{i})}{p(t_{j})} \cdot \frac{\pi(t_{j}\mid \Delta(x_{i}))}{\pi(t_{i}\mid \Delta(x_{i}))}
\end{equation}

\paragraph{Imbalance Loss.}
Based on the definition of the adjustment factor \( \Upsilon \), the distribution of learned adjustment factors \( \Upsilon(x) \) should be similar across treatment groups \citep{hassanpour2019learning}. To quantify the distributional differences of \( \Upsilon(x) \), \citet{hassanpour2019learning} utilize the Wasserstein distance, incorporating it as a penalty term in the loss function. We now extend this formulation to the multi-treatment case:  
\begin{equation}
 L_{Im}=\sum_{t=1}^{|T|} disc(\left \{ \Upsilon(x_{i})\right\}_{i:t_{i}=0} ,\left \{ \Upsilon(x_{i})\right\}_{i:t_{i}=t})
\end{equation}

\paragraph{Cross Entropy Loss.}
The combined representations of instrument factors and confounding factors contain sufficient information to reconstruct the decision rules for the treatment assignment \citep{hassanpour2019learning}. Therefore, \citet{hassanpour2019learning} leverage the learned representations $\Gamma(x)$ and $\Delta(x)$ to predict the assigned treatments, minimizing the loss function to ensure that all relevant information pertaining to treatment assignments is captured in these two representations. The following formula defines the loss function in the multi-treatment case:
\begin{equation}
    \pi(t|\Gamma(x),\Delta(x)) =\frac{e^{\varphi(\Gamma(x) ,\Delta(x) ) }_{t}}{\sum_{t=0}^{|T|}e^{\varphi(\Gamma(x) ,\Delta(x) ) }_{t}} 
\end{equation}

\begin{equation}
    L_{CE}=\frac{1}{N}\sum_{i=1}^{N}-\log \left[\pi \left( t_{i}|\Gamma(x_{i}),\Delta(x_{i}) \right) \right]
\end{equation}

\paragraph{Multi-task Learning Architecture for Estimating Factual Loss.}
The combination of the learned representations \( \Delta(x) \) and \( \Upsilon(x) \) retains sufficient information for accurate factual estimation. In scenarios with a finite number of treatments, drawing inspiration from the TarNet framework \citep{shalit2017estimating}, we employ a multi-task learning structure for factual estimation using \( \Delta(x) \) and \( \Upsilon(x) \). Specifically, each treatment \( t \) corresponds to a separate task network \( h^t(\cdot) \). We first obtain a shared representation for all tasks by passing the union of \( \Delta(x) \) and \( \Upsilon(x) \) through a network. Subsequently, each task network \( h^t(\cdot) \) is trained independently, allowing the model to learn treatment-specific responses while sharing common features across tasks. This multi-task approach enhances learning efficiency by leveraging shared information across treatments while capturing the nuances of each treatment assignment. When the outcome \( Y \) is binary, the factual loss function is calculated as follows:
\begin{equation}
    L_{Fac}(x, y, t) = - \left[ y \log \left( g^t(\Delta(x), \Upsilon(x)) \right) + (1 - y) \log \left( 1 - g^t(\Delta(x), \Upsilon(x)) \right) \right]
\end{equation}

\paragraph{ATE Loss.}
In practical contexts, such as optimizing the ROI for marketing campaigns, the accuracy of the average treatment effects is as crucial as the ranking of treatment effects. To enhance the estimate of average treatment effects, we employ the ATE loss, which is as follows:
\begin{equation}
    L_{ATE}=\frac{1}{|T|}\sum_{t=1}^{|T|} |\hat{\text{ATE}}_t - \text{ATE}_t| = \frac{1}{N \cdot |T|}\sum_{t=1}^{|T|} | (\sum_{i=1}^{N}\hat{y}_{i:t_i=t}-\sum_{i=1}^{N}\hat{y}_{i:t_i=0}) - (\sum_{i=1}^{N}y_{i:t_i=t}-\sum_{i=1}^{N}y_{i:t_i=0})|
\end{equation}

\paragraph{Monotonicity Penalty Loss.}
In online scenarios, such as modeling user purchase propensity at varying discount levels, the monotonicity assumption is often regarded as prior knowledge. As the treatment level increases, the individual response is expected to increase. To incorporate this knowledge into our model, we define the concept of monotonicity penalty loss, which penalizes any violation of this monotonicity assumption. The loss function is as follows:
\begin{equation}    L_{Mono}=\frac{1}{N}\sum_{i=1}^{N}\sum_{t=1}^{|T|}\text{ReLU}(\hat{y}_i^{t-1}-\hat{y}_i^{t})
\end{equation}

Here, \( \text{ReLU} \) denotes the rectified linear unit function, which ensures that the penalty is applied only when the predicted response at the higher treatment level is less than the smaller.

\paragraph{Overall Loss Function of DR-CFR}
Here, we review the above loss functions and present the overall loss function for the base causal effect learning model, defined as:
\begin{equation}
\begin{aligned}
L_{Base}&=\frac{1}{N}\sum_{i=1}^{N}w(t_{i},\Delta(x_{i}))\cdot L_{Fac}(x_i,y_i,t_i)\\
&+\alpha \cdot L_{CE} + \beta \cdot L_{Im} + \gamma \cdot L_{ATE} + \delta \cdot L_{Mono} + \lambda \cdot L_{Reg}\\
\end{aligned}
\end{equation}

where \( L_{Reg} \) is the regularization term, \( \alpha \), \( \beta \), \( \gamma \), \( \delta \) and \( \lambda \) are hyperparameters that control the scale of each component.

\subsection{The Continual Incremental Causal Effect Learning}
Building on the base causal effect learning model, we propose the ICE-PKD framework. This framework effectively addresses both covariate shift and concept drift by first training a well-generalized initial model on the initially available data and then updating it with incrementally available data.

\paragraph{Initial Model Selection and Parameter Initialization.}
Given the complexity of the uplift model, random initialization of the parameters can lead to significant performance perturbations. To mitigate this, we train the model multiple times on the initially available data and select the optimal model \( g_{\theta_0}^t(x) \). Furthermore, similar to the incremental learning approaches in computer vision \citep{li2017learning,rebuffi2017icarl}, we initialize the parameters of the incremental model based on those of the previously trained model. Specifically, during the $k$-th incremental training of the model, we initialize the model parameters \( \theta_{k} \) using \( \theta_{k-1} \), which are the parameters obtained after the $(k-1)$-th incremental training.

\paragraph{Loss Function Regularization via Knowledge Distillation.}
To ensure model generalization during incremental training, we draw inspiration from the class-incremental method LwF \citep{li2017learning} and adapt it to the uplift model scenario. Under the assumption of concept drift, our focus is the difference in predicted distributions between the model before and after incremental training. Specifically, we examine the discrepancy between the model \(g_{\theta_k}(x) \) and the model \( g_{\theta_{k-1}}(x) \). To quantify this difference, we employ the Kullback-Leibler (KL) Divergence. In the context of a binary outcome problem, the KL Divergence is computed as:
\begin{equation}
    D_\mathrm{KL}(y\parallel\hat{y})=y\log\frac{y}{\hat{y}}+(1-y)\log\frac{1-y}{1-\hat{y}}    
\end{equation}

where \( y \) represents the ground truth label and \( \hat{y} \) represents the predicted probability.

Thus, the knowledge distillation loss function $L_{KD}$ during the $k$-th incremental training is calculated as:
\begin{equation}
\label{eq:kd}
    L_{KD}=\frac{1}{N} \sum_{i=1}^{N} \sum_{t=0}^{|T|} D_\mathrm{KL}(g_{\theta_{k-1}}^{t}(x_i)\parallel{g_{\theta_{k}}^{t
}(x_i)})  
\end{equation}

where the superscript \( t \) indicates the model's response under the treatment \( t \), \( \theta_{k-1} \) are the model parameters after the $(k-1)$-th incremental training (which remain fixed during this process), and \( \theta_{k} \) are the model parameters updated during the current incremental training.

By penalizing the difference in predictive distributions between the model before and after incremental training, we aim to preserve previous knowledge and mitigate catastrophic forgetting.

\paragraph{Replay Mechanism.}
Due to covariate  shift, relying solely on new data to calculate distributional differences for preserving previous knowledge becomes unfeasible, as neural networks often perform poorly when predicting out-of-distribution data. Drawing inspiration from the class-incremental method iCaRL \citep{rebuffi2017icarl} in computer vision, we apply knowledge distillation to a replayed subset of the previous dataset. However, unlike image classification tasks, sample or feature centers do not play a significant role in our scenario. Therefore, we adopt a uniform sampling approach, preserving approximately one percent of the initial available sample size from the previous dataset as a representation of previous training data. Specifically, we set:
\begin{equation}
    N_{\text{save}}=\left\lfloor 0.01 \cdot \tilde{N}_0 \right\rfloor
\end{equation}

where \( \tilde{N}_0 \) is the sample size used for training the base model, and \( N_{\text{save}} \) represents the integer part of \( 0.01 \cdot \tilde{N}_0 \).

After the $(k-1)$-th incremental training, we have not only model parameters \( \theta_{k-1} \) but also the replayed dataset \( R_{k-1} \), which consists of \( N_{\text{save}} \) samples, along with the sample size \( \tilde{N}_{k-1} \) used for previous training. During the $k$-th incremental training, the latest available dataset \( D_{k} \) with \( N_{k} \) samples is utilized. The sample size used for previous training is then updated as follows:
\begin{equation}
    \tilde{N}_{k} = \tilde{N}_{k-1} + N_{k}
\end{equation}

To determine the size of the replayed dataset \( R_{k} \) from  \( R_{k-1} \) and \( D_{k} \), we calculate:
\begin{equation}
    N_{\text{save}}^R, N_{\text{save}}^D = \left\lfloor \frac{N_{\text{save}} \cdot \tilde{N}_{k-1}}{\tilde{N}_{k}} \right\rfloor, \left\lceil \frac{N_{\text{save}} \cdot N_k}{\tilde{N}_{k}} \right\rceil
\end{equation}

For uniform sampling, we generate sequences of random numbers:
\begin{equation}
    r_1,r_2,\ldots,r_{N_{\text{save}}},d_1,d_2,\dots,d_{N_{k}} \sim U(0,1)
\end{equation}

where \( \{ r_i \}_{i=1}^{N_{\text{save}}} \) and \( \{ d_i \}_{i=1}^{N_k} \) are sequences of random numbers corresponding to the indices of the datasets \( R_{k-1} \) and \( D_k \), respectively. These random sequences are sorted, and we select the first \( N_{\text{save}}^R \) samples from \( R_{k-1} \) and \( N_{\text{save}}^D \) samples from \( D_k \) to form the replayed dataset \( R_k \) for the $(k+1)$-th incremental training.

This sampling procedure can be expressed as:
\[
R_k \gets \text{UniformSample}(R_{k-1}, N_{\text{save}}^R) \cup \text{UniformSample}(D_k, N_{\text{save}}^D)
\]

\paragraph{Proxy Teacher for Knowledge Distillation.}
Concept drift introduces a natural discrepancy in the output distributions of the teacher model used for knowledge distillation, even without considering training bias. Consequently, the teacher model can produce a biased estimate of the label distribution. Requiring the student to blindly imitate the teacher model \( g_{\theta_{k-1}}(x) \) can lead to the student inheriting biases, resulting in suboptimal predictions during the $k$-th incremental training. To mitigate this issue, we transform the original teacher into a proxy teacher by incorporating a corrector \( h_{\theta_c}(x) \) in the calculation of KL divergence, drawing inspiration from the alternative solution of PTLoss \citep{zhang2024knowledge}. The loss function for knowledge distillation is then calculated as follows:
\begin{equation}
    L_{PT}=\frac{1}{N} \sum_{i=1}^{N} \sum_{t=0}^{|T|} D_\mathrm{KL}(g_{\theta_{k-1}}^{t}(x_i)+h_{\theta_c}^{t}(x_i)\parallel{g_{\theta_{k}}^{t}(x_i,t)})  
\end{equation}

This formulation is similar to Eq. \ref{eq:kd}. The difference is the addition of the corrector \( h_{\theta_c} \), which is a neural network sharing the same input covariates and output dimensions as \( g_{\theta_{k-1}}(x) \).

Additionally, we configure the activation function of the hidden layers to ReLU and the output layer to Tanh, while employing the Xavier initialization method to ensure that the output values of the \( h_{\theta_c} \) remain small before training. This mitigates large perturbations to the teacher model at the onset of training, facilitating more rapid convergence, which aligns well with incremental learning approaches that employ a smaller learning rate and fewer training epochs.

With this modification, not only are the updates to the model's parameters \( \theta_{k} \) influenced by the previous knowledge (i.e. teacher model \(g_{\theta_{k-1}}(x)\)), but also the teacher model is guided and corrected by the new data during the $k$-th incremental training. 

\paragraph{Overall Loss Function of ICE-PKD.}
The overall loss function of ICE-PKD during the $k$-th incremental training is calculated as follows:
\begin{equation}
    L_{ICE-PKD} = L_{Base}|D_{k} + \mu \cdot L_{PT}|R_{k-1}
\end{equation}

where \( \mu \) is a weighting coefficient, \( L_{Base}|D_{k} \) denotes \( L_{Base} \) computed on the latest available dataset \( D_{k} \), and \(  L_{PT}|R_{k-1} \) represents \( L_{PT} \) calculated on the replayed dataset \( R_{k-1} \).

Finally, we summarize the procedures of ICE-PKD. The pseudo code of ICE-PKD can be found at Algorithm \ref{alg:ice-pkd}.

\begin{algorithm}[htbp]
\caption{Incremental Causal Effect with Proxy Knowledge Distillation (ICE-PKD)}
\label{alg:ice-pkd}
\begin{algorithmic}[1]
\REQUIRE Initial dataset $D_0$, incremental data sequence $\{D_k\}_{k=1}^{|K|}$
\ENSURE Trained model $\theta_k$ for $k=1,\cdots,|K|$

\STATE \textbf{Initialize:}
\STATE Train base model: $\theta_0 \gets \arg\min_\theta L_{Base}|D_0$
\STATE Initialize replay buffer: $R_0 \gets \text{UniformSample}(D_0, N_{\text{save}})$
\STATE Set total sample count: $\tilde{N}_0 \gets N_0$

\FOR{$k=1$ \TO $|K|$}
    \STATE $\theta_k \gets \theta_{k-1}$ \COMMENT{Inherit previous parameters}
    
    \STATE Initialize corrector network $h_{\theta_c}$ using Xavier initialization
    \STATE Build proxy teacher: $g_{\theta_{k-1}}^{t}(x) + h_{\theta_c}^{t}(x)$
    
    \STATE Merge train datasets: $D_k \cup R_{k-1}$
    
    \FOR{epoch $=1$ \TO epochs}
        \STATE Update parameters: $\theta_k \gets \theta_k - \eta\nabla_{\theta_k}(L_{Base}|D_k + \mu \cdot L_{PT}|R_{k-1})$
        \STATE Update corrector: $\theta_c \gets \theta_c - \eta\nabla_{\theta_c}(L_{PT}|R_{k-1})$
    \ENDFOR

    \STATE Update total sample count: $\tilde{N}_k \gets \tilde{N}_{k-1} + N_k$
    \STATE Compute sample sizes: 
    \STATE $\quad N_{\text{save}}^R \gets \lfloor N_{\text{save}} \cdot \tilde{N}_{k-1} /\ \tilde{N}_{k} \rfloor$
    \STATE $\quad N_{\text{save}}^D \gets \lceil N_{\text{save}} \cdot N_{k} /\ \tilde{N}_{k} \rceil$
    \STATE $R_k \gets \text{UniformSample}(R_{k-1}, N_{\text{save}}^R) \cup \text{UniformSample}(D_k, N_{\text{save}}^D)$
\ENDFOR

\RETURN $\theta_k$ for $k=1,\cdots,|K|$
\end{algorithmic}
\end{algorithm}

\section{Experiments and Evaluations}

\subsection{Description of the datasets}
To accurately assess the model performance, we conduct experiments on both synthetic datasets and online datasets. The synthetic datasets \citep{berrevoets2019causal} align with our assumptions and provide access to ground truth (i.e., counterfactual outcomes). The online dataset further demonstrates the practical applicability and significance of the proposed method. The datasets are described as follows:

\subsubsection{Synthetic Dataset}
We first consider a 2-dimensional synthetic dataset \citep{berrevoets2019causal} with three treatments, facilitating both analysis and visualization. For each time period \( k \), the exact responses under different treatments are given by the following equations:
\begin{align}
    \label{eq:p0}
    p_{k,0}(x_1,x_2) &= \sigma\left(0.6x_1^2 + x_1x_2^2 - (0.5-d_0(k))\cdot x_1 - (0.6 - 0.5d_0(k)) \cdot x_2 + 0.2d_0(k) + \epsilon \right) \\
    \label{eq:p1}
    p_{k,1}(x_1,x_2) &= \sigma\left(0.7x_1^2 + 0.1x_2^2 + x_1x_2^2 - (0.5-d_1(k))\cdot x_1 - (0.5 - 0.5d_1(k)) \cdot x_2 + 0.2d_1(k) + \epsilon \right) \\
    \label{eq:p2}
    p_{k,2}(x_1,x_2) &= \sigma\left(0.9x_1^2 + 0.15x_2^2 + x_1x_2^2 - (0.5-d_2(k))\cdot x_1 - (0.5 - 0.5d_2(k)) \cdot x_2 + 0.2d_2(k) + \epsilon \right)
\end{align}

where \( x_1,x_2 \in [0,1] \), and \( \sigma( \cdot ) \) represent the logistic sigmoid function, which maps the polynomial expressions to the range \( [0,1] \). The term \( \epsilon \sim \mathcal{N}(0, 0.0001) \) represents Gaussian noise, and \( x = (x_1, x_2) \) is a covariate with two components. The treatments \( t=0,1,2 \) correspond to different coupon amounts, while \( d_0(k), d_1(k), d_2(k) \) represent concept drift that varies over time, indexed by the time period \( k \). The specific relationship between \(d_i(k)\) and \( k \) is defined as follows:
\begin{equation}
    d_i(k)=\begin{cases} 
0.1ik + 0.2k & \text{if } 0 \leq k \leq 3 \\
-0.1ik + 0.65i - 0.2k + 1.3 & \text{if } 4 \leq k \leq 6 
\end{cases}
\quad \text{for } i = 0, 1, 2
\end{equation}

\begin{figure}
    \centering
    \includegraphics[height=6cm]{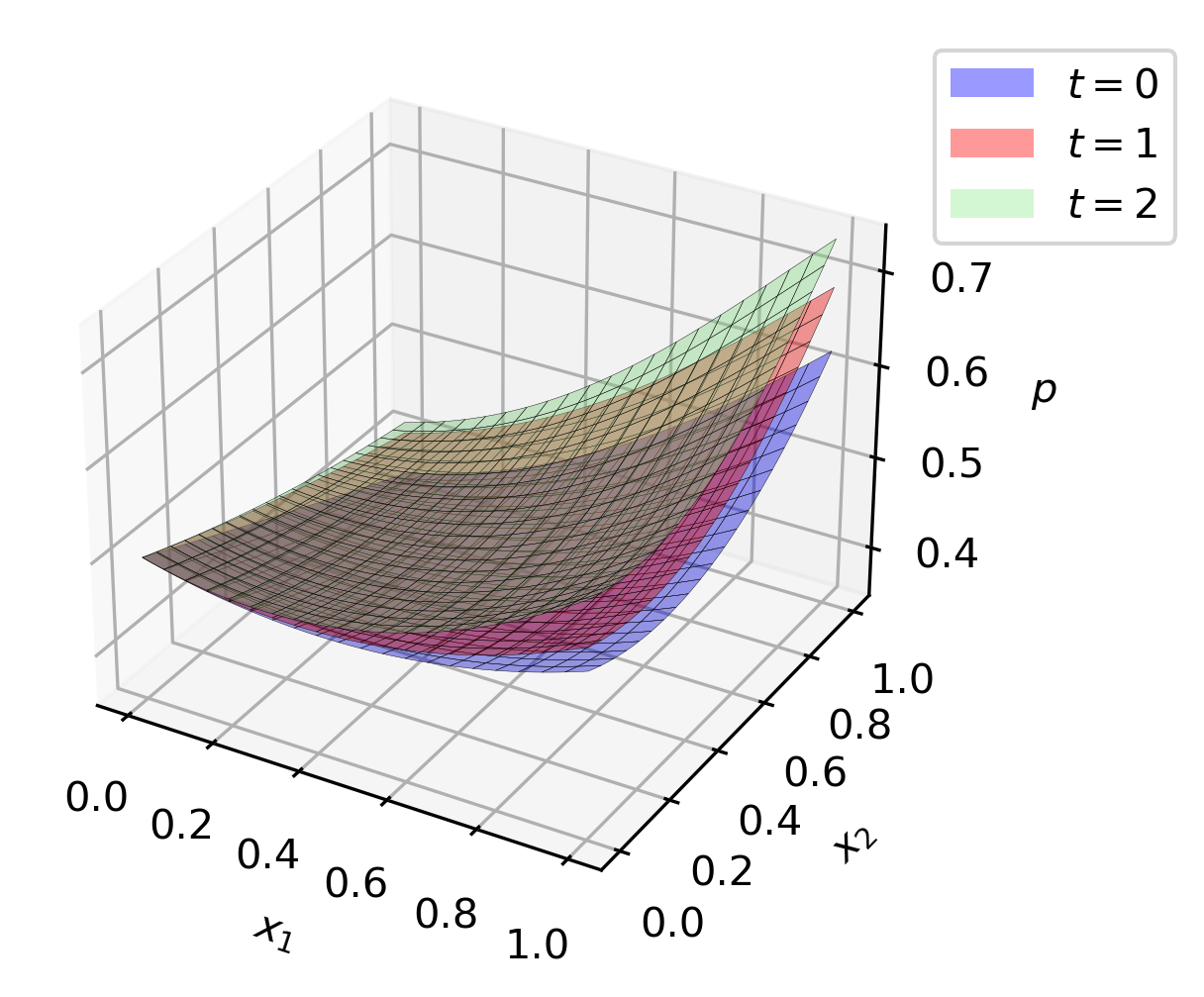} 
    \caption{Response Function Visualization for \(k=0\). The blue surface corresponds to \(t=0\), the red surface to \(t=1\), and the green surface to \(t=2\).}
    \label{fig:fig2}
\end{figure}

Figure~\ref{fig:fig2} presents the visualization of the response function for the time period \( k=0 \). As shown, the generated functional relationships adhere to the monotonicity assumption, aligning with online expectations. Specifically, a user's willingness to purchase increases as the discount rises.

To satisfy the assumption of covariate shift, we generate a training covariate dataset of size 10,000 and a test covariate dataset of size 50,000, utilizing four truncated normal distributions and one uniform distribution for each time period \( k \). The means of the truncated normal distributions are randomly selected from the range \( [0.2,0.8] \), while their standard deviations are chosen from \( [0.01,0.1] \), with covariance set to zero. According to the properties of normal distributions, most points lie within the domain \( [0,1] \times [0,1] \), and any points outside this domain are truncated to fall within the domain boundaries. The uniform distribution is denoted as \( U[0,1] \). 

According to Eqs. \ref{eq:p0}, \ref{eq:p1} and \ref{eq:p2}, we can derive the exact response probabilities for the three treatments, which are subsequently used to evaluate the Precision in Estimation of Heterogeneous Effect (PEHE). Additionally, these response probabilities serve as parameters in Bernoulli experiments to generate the corresponding labels. 

In the test dataset, each covariate is randomly assigned a treatment, ensuring the absence of selection bias. In the training dataset, treatments are applied according to predefined rules. Specifically, when \( x_1 + x_2 < 0.8\), neither the uplift from \( t=0 \) to \( t=1 \) nor from \( t=1 \) to \( t=2 \) is significant. In this case, we prefer to apply treatment \( t=0 \) to covariates in this region, while the probabilities of applying \( t=1 \) or \( t=2 \)  are both relatively low. Practically, we perform a Bernoulli test with a probability of 0.6 and apply \( t=0 \) if successful; otherwise, we conduct another Bernoulli test with a probability of 0.5 and apply \( t=1 \) if successful. If neither test is successful, we apply \( t=2 \). Conversely, when \( x_1 + x_2 > 1.2 \), treatment \( t=1 \) is preferred, while for \( 0.8 \leq x_1 + x_2 \leq 1.2 \), treatment \( t=2 \) is preferred.

For the 10-dimensional synthetic dataset, we replicate the process used for the 2-dimensional dataset. The primary distinction is the increased dimensionality, which results in a larger dataset for both training and testing.

\subsubsection{Online Dataset}
In Huaxiaozhu's new passenger acquisition campaign, we have access to substantial data that exhibits both covariate shift and conceptual drift. To capture these dynamics, we partition the data over time, following the model updating strategy employed in previous online work. Specifically, the non-incremental model updating strategy utilizes approximately 1 million records from the past 9 weeks each week, balancing resource constraints with model effectiveness. For each time period \( k \), the records from the subsequent week are considered as the test dataset. We use around 120,000 new passenger records each week as an incremental training set \( D_k \) for time period \( k=1,2,\cdots,6 \), while the initially available dataset comprises approximately 1 million records from the 9 weeks preceding the start of the experiment. 

For modeling, we utilize a 53-dimensional feature space that includes covariates such as city-specific characteristics and historical statistics. The campaign involves eight treatment levels, corresponding to coupon amounts \( 3,4, \dots,10 \), with treatments indexed as \( t=0,1, \dots, 7 \). The control group is represented by \( t=0 \), which corresponds to a minimum coupon amount of \( 3 \). The outcome label \( y \in \{ 0,1 \} \) indicates whether an order is completed (\( y=0 \)) or not (\( y=1 \)). 

\subsection{Experimental Setting}
\label{section:experimental setting}
For the synthetic dataset, we sequentially split the 10\% of each training dataset for the time period \( k \) as the available dataset \( D_k \). This criterion is motivated by the observed decline in online performance during this period. To predict the test dataset for each time period \( k \), where \( k=1,2,\cdots,6\), we apply our model, ICE-PKD, along with three comparison strategies. In the incremental updating of ICE-PKD, we first select an initial model \( g_{\theta_0}^t(x) \) that achieves optimal performance on the initially available data (i.e. the entire training dataset at the time period \( k=0 \)) and then incrementally update the model with \( D_k \) for each \( k=1,2,\cdots,6 \). To further demonstrate the effectiveness of ICE-PKD, we compare it with three alternative strategies: DR-CFR A trains the DR-CFR using the entire training dataset from the previous time period \( k-1 \), serving as the baseline; DR-CFR B trains the DR-CFR using only the latest available dataset \( D_k \), which limits generalization; DR-CFR C does not update the model \( g_{\theta_0}^t(x) \), sacrificing real-time adaptability. 

For the online dataset, we simulate a scenario where the exact time of drift occurrence is unknown, assuming that drift emerges gradually over time. To predict the test dataset for each time period \( k \), where \( k=1,2,\cdots,6\), we again apply ICE-PKD and three comparison strategies. In the incremental updating of ICE-PKD, we select an initial model \( g_{\theta_0}^t(x) \) that performs best on the initially available records and then update the model incrementally with \( D_k \). DR-CFR A trains the DR-CFR using the records from the past 9 weeks (including \( D_k \)), serving as the baseline. DR-CFR B trains the DR-CFR using only the latest available dataset \( D_k \). DR-CFR C does not update the model \( g_{\theta_0}^t(x) \). Table \ref{tab:strategy_comparison} presents a comparative analysis of various strategies, focusing on key aspects such as the use of the latest data, data volume, and the application of incremental updates. 

Throughout the experiment, all settings remain consistent, except for the configuration of ICE-PKD, which introduces an additional corrector and utilizes a smaller learning rate along with fewer training epochs. Specifically, for the 2-dimensional synthetic dataset, ICE-PKD has 8,952 parameters, while the other non-incremental models have 8,409 parameters. The learning rate is set to $1\times10^{-4}$ for ICE-PKD and $1\times10^{-3}$ for the others. Training epochs are set to 10 for ICE-PKD and 20 for the others. For the 10-dimensional synthetic dataset, ICE-PKD has 50,372 parameters, while the others have 48,169. For the online dataset, ICE-PKD has 463,612 parameters, while the others have 456,204.

\begin{table}
	\caption{Comparative Analysis of Various Strategies}
	\centering
	\begin{tabular}{lcccc}
		\toprule
		\multicolumn{1}{c}{\textbf{Strategy}} & \multicolumn{1}{c}{\textbf{Use of Latest Data}} & \multicolumn{1}{c}{\textbf{Data volume}} & \multicolumn{1}{c}{\textbf{Incremental Updates}} \\
		\midrule
		DR-CFR A & Yes & $100,000$  & No \\
		DR-CFR B & Yes & $14,000$  & No \\
		DR-CFR C & No & $100,000$  & No \\
		ICE-PKD & Yes & $100,000+k\times14,000$  & Yes \\
		\bottomrule
	\end{tabular}
	\label{tab:strategy_comparison}
\end{table}

\subsection{Evaluation Metrics}
In this study, we adopt three commonly used evaluation metrics: Average Treatment Effect Error (ATE Error), Precision in Estimation of Heterogeneous Effect (PEHE) \citep{hill2011bayesian} and QINI Coefficient \citep{devriendt2020learning}. These metrics are essential for assessing the performance of model predictions. However, a gap (constrained optimization method) exists between model predictions and online  decision-making. Therefore, improved predictions do not necessarily translate to increased online revenues. To address this, we introduce a novel simulation metric, Area Under Cost Curve for Resource Allocation Simulation (RAS-AUCC), which provides a more accurate reflection of the overall performance of online strategies using randomized controlled trial(RCT) data. The specific formulas and principles are as follows:

\paragraph{ATE Error.}
ATE Error quantifies the difference between the estimated and true average treatment effects for the treatment group. It helps evaluate the accuracy in predicting the overall treatment effect. In scenarios with multiple treatments, the ATE Error is calculated as the average across all treatment groups. The formula is given by: 
\begin{equation}
   \epsilon_{ATE} = \frac{1}{|T|}\sum_{t=1}^{|T|}|\hat{\text{ATE}}_t - \text{ATE}_t| 
\end{equation} 

where \( \hat{\text{ATE}}_t\) is the estimated ATE for the treatment group \( t \) and \( \text{ATE}_t \) is the true ATE for the treatment group \( t \).
 
\paragraph{PEHE.}
PEHE evaluates the model's accuracy in predicting heterogeneous treatment effects across the entire population, specifically applied to synthetic datasets. This metric is computed as the square root of the average squared differences between the predicted and ground truth treatment effects, weighted by the population size. The formula of PEHE is: 
\begin{equation}
   \epsilon_{PEHE} = \frac{1}{N \cdot |T|}\sum_{i=1}^{N}\sum_{t=1}^{|T|}{(\hat{\tau}_i^t - \tau_i^t)}^2
\end{equation}

where \( \hat{\tau}_i^t \) is the predicted treatment effect for individual \( i \) in treatment group \( t \), \( \tau_i^t \) is the true treatment effect for individual \( i \) in treatment group \( t \) and \( N \) is the total number of individuals.

\paragraph{QINI.}

The QINI Coefficient is a metric designed to evaluate the ranking performance of uplift estimation. It quantifies the difference in rewards between treatment and control groups on a per-segment basis, following the sorting of samples by their estimated uplift. To enhance its applicability in multi-treatment scenarios, we employ the average QINI Coefficient as a performance measure. The QINI Coefficient is calculated as the area under the QINI Curve. Sort all the quintuples from the treatment \( \tilde{t} \) and control groups in descending order of estimated individual treatment effect. Let \( m \) represent a percentage (e.g. 10\%, 20\%) and define \( S_m \) as the set of samples corresponding to the top \( m \) percent. For treatment \( \tilde{t} \), the values for the QINI Curve are computed as follows:

\begin{equation}
    V(m) = \sum_{i \in S_m} \mathbb{I}(y_i=1, t_i=\tilde{t}) - \frac{\sum_{i \in S_m} \mathbb{I}(y_i=1, t_i=0) \sum_{i \in S_m} \mathbb{I}(t=\tilde{t})}{\sum_{i \in S_m} \mathbb{I}(t_i=0)} 
\end{equation}

where \( \mathbb{I}(\cdot) \) denotes the indicator function, such that \( \mathbb{I}(t_i=0) \) equals 1 when \( t_i=0\) and 0 otherwise.

\paragraph{RAS-AUCC.} We assume the availability of a RCT dataset consisting of \( N \) samples. Given a specific budget \( b \), we can determine the decision-making allocation \( t_i^b \) based on the model predictions for each individual \( i \). The allocation method commonly employed is a constrained optimization algorithm in marketing, such as 0-1 integer programming and greedy algorithms. In the RCT dataset, we can obtain the ground truth cost and reward associated with a treatment value for each individual \( i \), denoted as \( c_i \) for cost and \( r_i \) for reward. This allows us to estimate the incremental cost \( \Delta Cost(b) \) and the incremental reward \( \Delta Reward(b) \).The specific formulas are as follows:
\begin{equation}
    \Delta Cost(b) = N \cdot \left( \frac{\sum_i \mathbb{I}(t_i > 0, t_i = t_i^b) c_i}{\sum_i \mathbb{I}(t_i > 0, t_i = t_i^b)} - \frac{\sum_i \mathbb{I}(t_i = 0) c_i}{\sum_i \mathbb{I}(t_i = 0)} \right)
\end{equation}

\begin{equation}
    \Delta Reward(b) = N \cdot \left( \frac{\sum_i \mathbb{I}(t_i > 0, t_i = t_i^b) r_i}{\sum_i \mathbb{I}(t_i > 0, t_i = t_i^b)} - \frac{\sum_i \mathbb{I}(t_i = 0) r_i}{\sum_i \mathbb{I}(t_i = 0)} \right)
\end{equation}

\begin{figure}
    \centering
    \includegraphics[height=6cm]{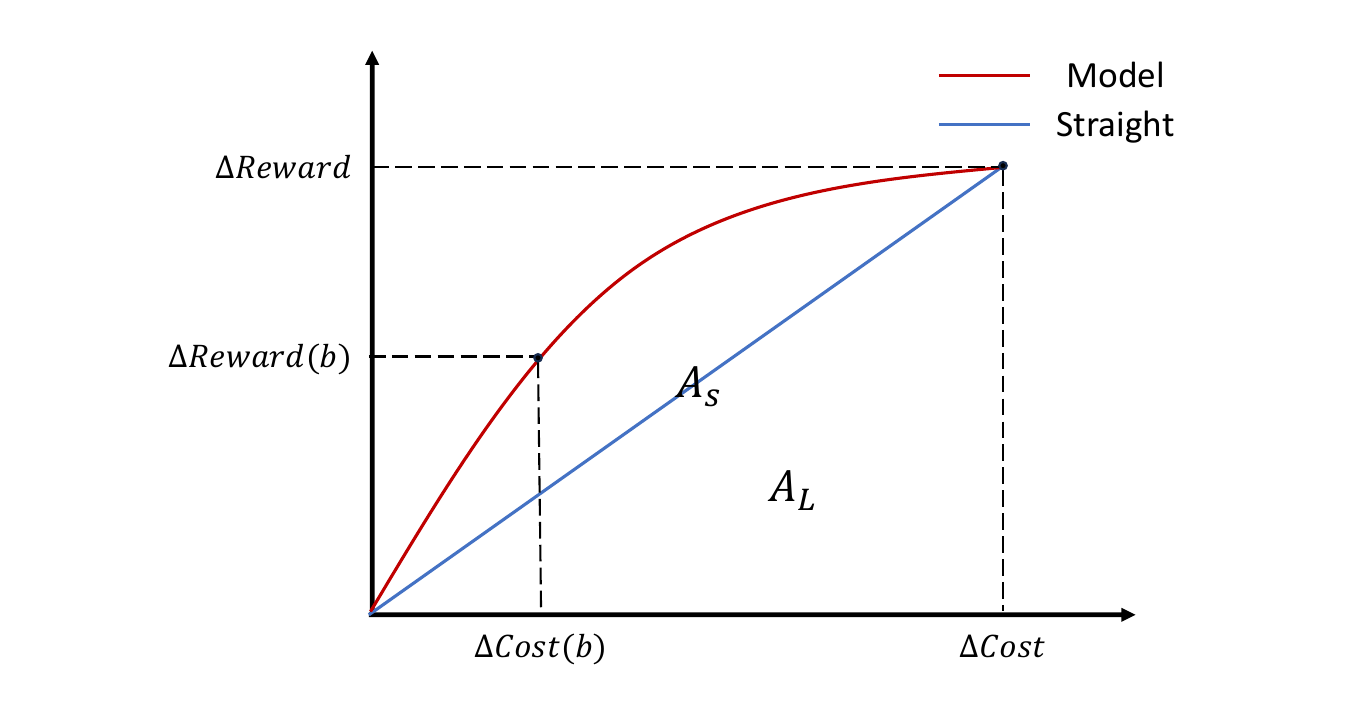} 
    \caption{RAS-AUCC. The red line represents the model's predicted curve, while the green line represents the straight-line benchmark. $\Delta Cost$ and $\Delta Reward$ indicate the maximum cost and the maximum reward when the largest treatment is applied to each individual.}
    \label{fig:RAS-AUCC}
\end{figure}

Our objective is to maximize the ratio of the incremental rewards to the incremental costs, i.e. ROI. Similar to the QINI curve, we can derive a curve based on the \( \Delta Cost(b) \) and \( \Delta Reward(b) \). As illustrated in Figure \ref{fig:RAS-AUCC}, the area under this curve is \(A_S\), while the area under the straight line is \(A_L\), which is used for normalization. Consequently, the RAS-AUCC is calculated as \((A_S - A_L)/A_L\).

\subsection{Evaluation Framework for Continual Causal Effects Estimation}
Estimating continual causal effects involves various evaluation metrics across multiple time periods, complicating the assessment of model performance. Therefore, establishing a comprehensive and robust assessment framework is crucial. A common approach is to compare the average differences of each metric across time periods. However, this method is highly sensitive to extreme values. Specifically, a significant increase or decrease in one time period can distort the observed effects in others. To address this challenge, we propose the use of two stability indicators to jointly evaluate the model's performance.

\paragraph{PRIO-10.}
The Percentage of Relative Improvement Over 10\% (PRIO-10) measures the proportion of cases where the model's improvement exceeds 10\%. This indicator assesses the overall performance gain relative to the baseline. Specifically, given the Baseline \( M_{\text{baseline}} \), for any model \( A \), and for any evaluation metric \( E \) that benefits from larger values (such as QINI and RAS-AUCC), the improvement \citep{zhong2022descn} is calculated as:
\begin{equation}
    Impr_k(A)=\frac{E_k(A)-E_k(M_{\text{baseline}})}{|E_k(M_{\text{baseline}})|} \times 100 \%, \quad k=1,2,\cdots,|K|
\end{equation}

where \(k\) represents the different time periods. The PRIO-10 for method \( A \) is then calculated as:
\begin{equation}
    \text{PRIO-10}(A)=\frac{1}{|K|}\sum_{k=1}^{|K|}\mathbb{I}(Impr_k(A)>10\%) \times 100\%
\end{equation}

In contrast, for any evaluation metric that improves with smaller values, such as ATE Error and PEHE, the improvement is calculated by multiplying the result by \( -1 \).

\paragraph{PRDU-5.}
The Percentage of Relative Deterioration Under 5\% (PRDU-5) measures the proportion of cases in which the model's performance deteriorates under 5\%. This metric evaluates the model's ability to avoid significant declines in performance, providing insight into its stability across different time periods.

\paragraph{AD.}
The Average Difference (AD) assesses the model's overall performance by calculating the average difference in evaluation metrics between the model and the baseline across all time periods. This indicator offers a straightforward measure of how well the model performs relative to the baseline.

\section{Experiment Results}
This section presents the results of our experiments, providing a comparative analysis of various strategies,and conducting an ablation experiment to explain the contribution of each module.

\subsection{Performance Comparison}
We assess the effectiveness of ICE-PKD by comparing it against other strategies on both synthetic and online datasets, as detailed in Table \ref{tab:comparison}. The experimental setup is described in Section \ref{section:experimental setting}. Key insights from the results are summarized below.

Compared to other strategies, ICE-PKD consistently demonstrates superior performance on both synthetic and online datasets. Specifically, ICE-PKD achieves the largest reductions in the error metrics \( \epsilon_{ATE} \) and \( \epsilon_{PEHE} \), as well as almost the most significant improvements in the order metrics QINI and RAS-AUCC. Notably, it is the only strategy that surpasses the baseline (DR-CFR A) across all indicators. Emphasize here, the robustness of the AD is evidenced by the performance of all other indicators, while the confidence of RAS-AUCC is reflected in other metrics.

In contrast, DR-CFR B, which is trained solely on the most recent dataset \( D_k \), demonstrates lower performance in order metrics. On the 2-dimensional synthetic dataset, it is outperformed by both DR-CFR C and ICE-PKD in terms of \( \epsilon_{PEHE} \). Although DR-CFR B performs better than DR-CFR C in AD for RAS-AUCC, it significantly underperforms in PRIO-10 and PRDU-5, which is  undesirable. On the 10-dimensional synthetic dataset, it performs worse than DR-CFR C in AD for RAS-AUCC, further underscoring its unreliability. On the online dataset, DR-CFR B's performance is even worse, significantly trailing the baseline in both QINI and RAS-AUCC. These results suggest that DR-CFR B lacks generalization and may overfit when trained on smaller datasets. 

DR-CFR C, which does not update the initial model \(g_{\theta_0}^t(x)\), exhibits poorer performance across the error metrics. Notably, it performs the worst in \( \epsilon_{ATE} \), being the only strategy significantly below the baseline for both the 2-dimensional synthetic dataset and the online dataset. Although DR-CFR C performs well on some AD indicators, its stability indicators (PRIO-10 and PRDU-5) are generally the worst, indicating it is not a reliable choice. This poor performance can be attributed to the model's lack of real-time adaptability in the presence of covariate shift and concept drift.

\begin{figure}[htbp]
    \centering
    \begin{minipage}{0.24\textwidth} 
        \centering
        \includegraphics[width=\textwidth]{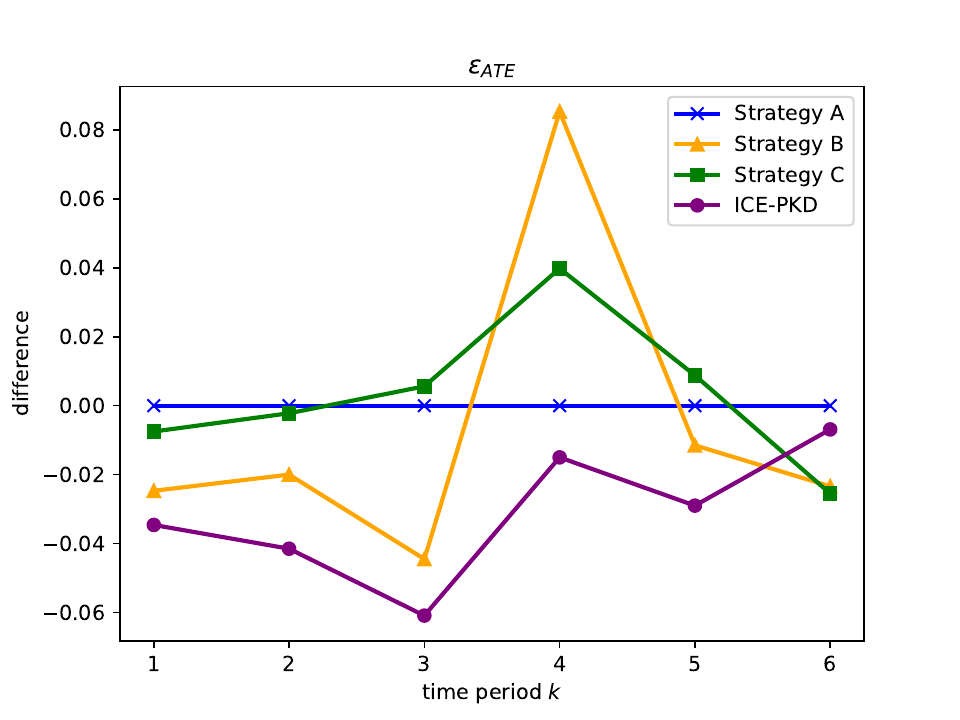}
    \end{minipage}\hfill
    \begin{minipage}{0.24\textwidth}
        \centering
        \includegraphics[width=\textwidth]{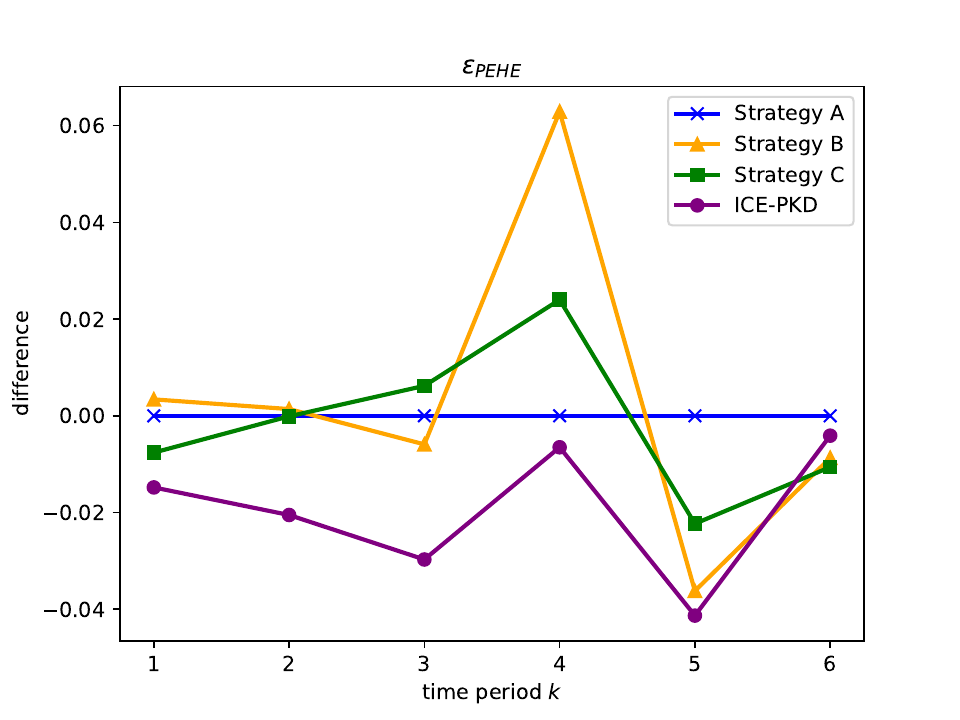}
    \end{minipage}\hfill
    \begin{minipage}{0.24\textwidth}
        \centering
        \includegraphics[width=\textwidth]{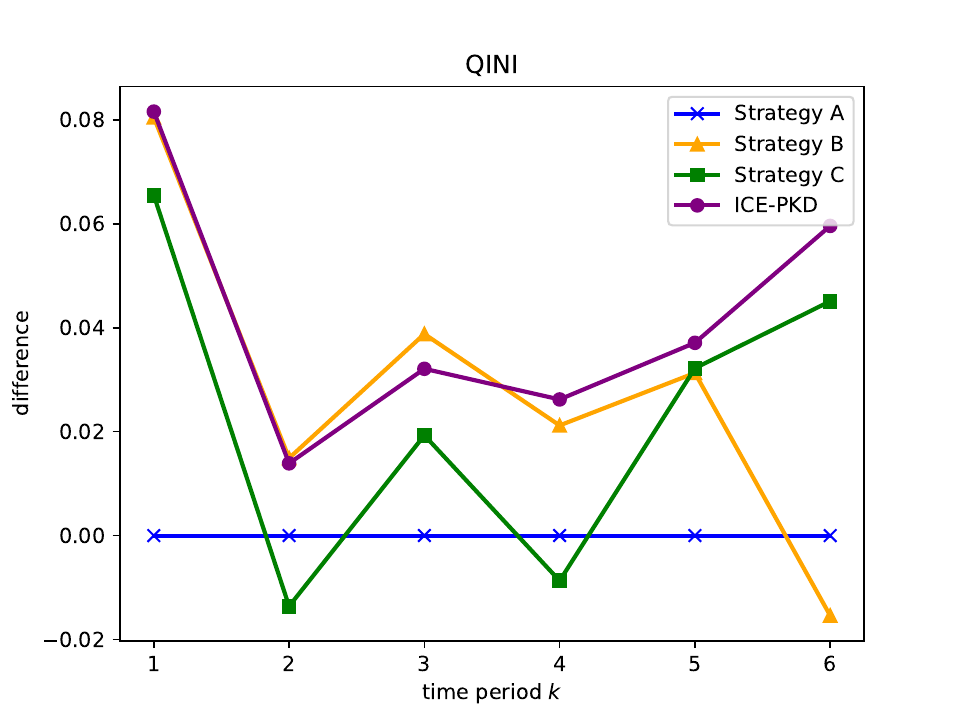}
    \end{minipage}\hfill
    \begin{minipage}{0.24\textwidth}
        \centering
        \includegraphics[width=\textwidth]{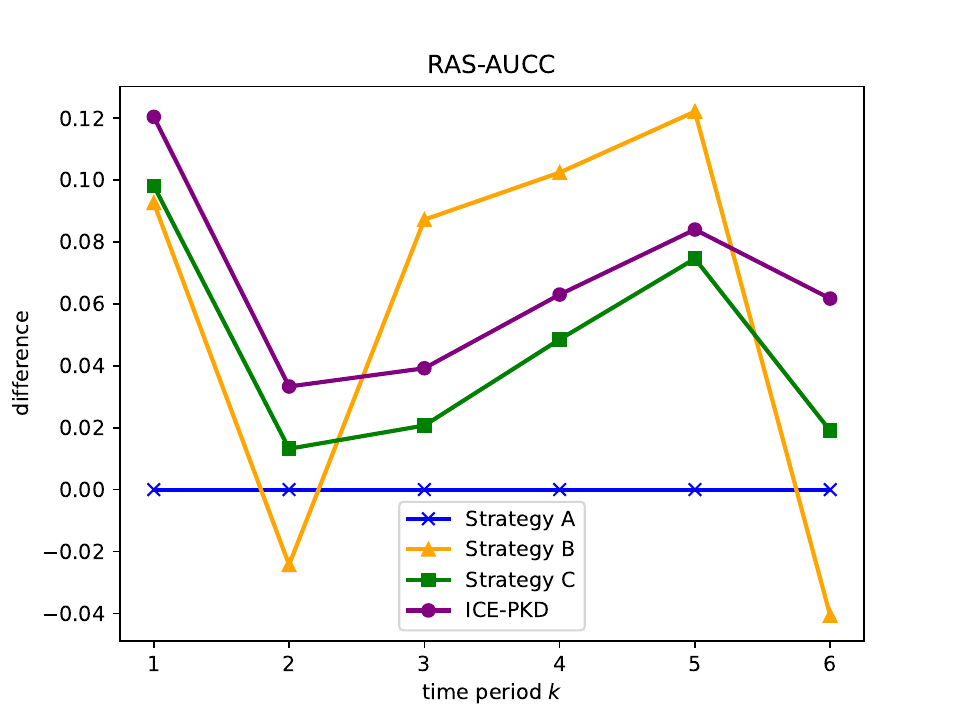}
    \end{minipage}
    \caption{Comparison of performance metrics across different strategies. Each subfigure illustrates a specific metric, with the horizontal axis representing different time periods \(k\) and the vertical axis depicting the difference in metric values between the various strategies and the baseline (DR-CFR A).}
    \label{fig:comparison}
\end{figure}

To further illustrate these findings, we visualize the comparison results for the 2-dimensional synthetic dataset in Figure \ref{fig:comparison}. The vertical axis represents the difference between each strategy's metrics and the baseline's across different time periods \( k \). ICE-PKD (represented by the purple line with circular marker points) consistently performs best, positioned at the bottom of the subplots for \( \epsilon_{ATE} \) and \( \epsilon_{PEHE} \), and at the top of the subplots for QINI and RAS-AUCC. In contrast, DR-CFR B (yellow line with triangle markers) exhibits significant fluctuations due to the smaller training set. DR-CFR C (green line with square markers) performs poorly, particularly in \( \epsilon_{ATE} \) and \( \epsilon_{PEHE} \), due to its lack of real-time adaptability. These visual observations closely align with the analytical results discussed above. 

In conclusion, ICE-PKD outperforms all other strategies across both synthetic and online datasets, demonstrating its robustness, adaptability and effectiveness in addressing covariate shift and concept drift.

\begin{table}
	\caption{Performance comparison of different strategies on synthetic and online datasets}
	\centering
        \resizebox{\textwidth}{!}{
	\begin{tabular}{clllllllllllll}
		\toprule
		& & \multicolumn{3}{c}{$\epsilon_{ATE}$} & \multicolumn{3}{c}{$\epsilon_{PEHE}$} & \multicolumn{3}{c}{QINI} & \multicolumn{3}{c}{RAS-AUCC}\\
		\cmidrule(lr){3-5} \cmidrule(lr){6-8} \cmidrule(lr){9-11} \cmidrule(lr){12-14}
            & strategy & \small{PRIO-10} & \small{PRDU-5} & \small{AD} & \small{PRIO-10} & \small{PRDU-5} & \small{AD} & \small{PRIO-10} & \small{PRDU-5} & \small{AD} & \small{PRIO-10} & \small{PRDU-5} & \small{AD}\\
            \midrule
            \multirow{3}{*}[0pt]{\parbox[c][3\baselineskip]{1cm}{\centering synthetic \newline  dataset \newline 2d}}
                & DR-CFR A & - & - & 0 & - & - & 0 & - & - & 0 & - & - & 0\\
                & DR-CFR B & 83.3\% & 83.3\% & -0.0065 & 50\% & 66.7\% & 0.0028 & 83.3\% & 83.3\% & 0.0286 & 66.7\% & 66.7\% & 0.0566\\
                & DR-CFR C & 33.3\% & 50\% & 0.0032 & 50\% & 66.7\% & -0.0017 & 66.7\% & 66.7\% & 0.0233 & 100\% & 100\% & 0.0457\\
                & ICE-PKD & 100\% & 100\% & -0.0313 & 100\% & 100\% & -0.0195 & 100\% & 100\% & 0.0418 & 100\% & 100\% & 0.0669\\
            \midrule
            \multirow{3}{*}[0pt]{\parbox[c][3\baselineskip]{1cm}{\centering synthetic \newline  dataset \newline 10d}}
                & DR-CFR A & - & - & 0 & - & - & 0 & - & - & 0 & - & - & 0\\
                & DR-CFR B & 16.7\% & 33.3\% & 0.0073 & 33.3\% & 66.7\% & -0.0164 & 16.7\% & 33.3\% & -0.0019 & 66.7\% & 66.7\% & 0.0161\\
                & DR-CFR C & 0\% & 0\% & 0.0154 & 33.3\% & 33.3\% & -0.0104 & 16.7\% & 16.7\% & -0.0155 & 66.7\% & 66.7\% & 0.0540\\
                & ICE-PKD & 83.3\% & 100\% & -0.0177 & 100\% & 100\% & -0.0276 & 100\% & 100\% & 0.0112 & 66.7\% & 100\% & 0.0699\\
            \midrule
            \multirow{3}{*}{\parbox[c]{1cm}{online\\ dataset}} 
                & DR-CFR A & - & - & 0 & - & - & - & - & - & 0 & - & - & 0\\
                & DR-CFR B & 50\% & 50\% & -0.0002 & - & - & - & 33.3\% & 33.3\% & -0.0289 & 50\% & 50\% & -0.0097\\
                & DR-CFR C & 33.3\% & 33.3\% & 0.0039 & - & - & - & 66.7\% & 83.3\% & 0.0213 & 83.3\% & 100\% & 0.0789\\
                & ICE-PKD & 66.7\% & 100\% & -0.0023 & - & - & - & 83.3\% & 83.3\% & 0.0273 & 83.3\% & 100\% & 0.0495\\
		\bottomrule
	\end{tabular}}
	\label{tab:comparison}
\end{table}

\subsection{Ablation Experiment}
To evaluate the contribution of each component in ICE-PKD, we conducted an ablation experiment on the 2-dimensional synthetic dataset. In this experiment, we systematically removed key components of the model to assess their impact on performance. The results are presented in Table \ref{tab:ablation}.

We define three variants of ICE-PKD by removing the following components:

w/o PT: ICE-PKD without the proxy teacher for knowledge distillation.

w/o RM: ICE-PKD without the replay mechanism.

w/o KD: ICE-PKD without the knowledge distillation process.

For each variant, the experimental setup is consistent with ICE-PKD. The results in Table \ref{tab:ablation} yield the following key insights:

\textbf{Impact of Proxy Teacher (PT):} The removal of the proxy teacher for knowledge distillation (w/o PT) results in a significant decline in overall performance across all metrics. Specifically, \( \epsilon_{ATE} \) and \( \epsilon_{PEHE} \) exhibit increased error values, with the AD for these metrics worsening from -0.0313 to -0.0270 and from -0.0195 to -0.0171, respectively. Additionally, both QINI and RAS-AUCC show decreased values, with the AD dropping from 0.0418 to 0.0397 and from 0.0669 to 0.0634, respectively. These results underscore the critical role of the proxy teacher in knowledge distillation, as it enhances the overall performance of ICE-PKD by preventing the model from blindly mimicking the teacher in the presence of concept drift.

\textbf{Impact of Replay Mechanism (RM):} The exclusion of the replay mechanism (w/o RM) leads to the most pronounced decrease in performance, especially for RAS-AUCC. Specifically, the PRIO-10 drops from 100\% to 83.3\%, and the AD decreases from 0.0669 to 0.0434. These findings highlight the importance of preserving previous knowledge through the replayed dataset. Given that neural network models often struggle with out-of-distribution data, relying solely on knowledge distillation with the latest available data can misguide the model when faced with covariate shift.

\textbf{Impact of Knowledge Distillation (KD):} The absence of the knowledge distillation process (w/o KD) significantly undermines stability indicators. Specifically, PRIO-10 for \( \epsilon_{ATE} \) and \( \epsilon_{PEHE} \) decreases to 83.3\% and 66.7\%, respectively, while PRDU-5 for \( \epsilon_{ATE} \) also drops to 83.3\%. These results illustrate the essential role of knowledge distillation in maintaining stable and accurate predictions over time.

\textbf{ICE-PKD:} The ICE-PKD framework consistently outperforms its ablated variants across all metrics, achieving the lowest average error metrics, the highest average order metrics, and the highest stability indicators. For instance, ICE-PKD achieves 100\% in both PRIO-10 and PRDU-5 for \( \epsilon_{PEHE} \) and RAS-AUCC, along with the lowest AD values for \( \epsilon_{PEHE} \) and the highest for RAS-AUCC. These results confirm the effectiveness of integrating all components in the proposed method.

In conclusion, the ablation experiment validates the design of ICE-PKD, highlighting the critical contributions of proxy teacher, replay mechanism and knowledge distillation to the overall performance. Collectively, these components enhance the model's robustness, stability and accuracy. 

\begin{table}
	\caption{Ablation study on synthetic dataset}
	\centering
        \resizebox{\textwidth}{!}{
	\begin{tabular}{clllllllllllll}
		\toprule
		& & \multicolumn{3}{c}{$\epsilon_{ATE}$} & \multicolumn{3}{c}{$\epsilon_{PEHE}$} & \multicolumn{3}{c}{QINI} & \multicolumn{3}{c}{RAS-AUCC}\\
		\cmidrule(lr){3-5} \cmidrule(lr){6-8} \cmidrule(lr){9-11} \cmidrule(lr){12-14}
            & strategy & \small{PRIO-10} & \small{PRDU-5} & \small{AD} & \small{PRIO-10} & \small{PRDU-5} & \small{AD} & \small{PRIO-10} & \small{PRDU-5} & \small{AD} & \small{PRIO-10} & \small{PRDU-5} & \small{AD}\\
            \midrule
            \multirow{3}{*}{\parbox[c]{1cm}{synthetic\\ dataset}} 
                & ICE-PKD & 100\% & 100\% & -0.0313 & 100\% & 100\% & -0.0195 & 100\% & 100\% & 0.0418 & 100\% & 100\% & 0.0669\\
                & w/o PT & 83.3\% & 100\% & -0.0270 & 83.3\% & 100\% & -0.0171 & 100\% & 100\% & 0.0397 & 100\% & 100\% & 0.0634\\
                & w/o RM & 100\% & 100\% & -0.0257 & 83.3\% & 100\% & -000169 & 100\% & 100\% & 0.0421 & 83.3\% & 100\% & 0.0434\\
                & w/o KD & 83.3\% & 83.3\% & -0.0271 & 66.7\% & 100\% & -0.0170 & 100\% & 100\% & 0.0407 & 100\% & 100\% & 0.0637\\
		\bottomrule
	\end{tabular}}
	\label{tab:ablation}
\end{table}

\subsection{Online Performance}
We have conducted an online A/B test experiment in Huaxiaozhu's new passenger acquisition campaign. By deploying the proposed incremental updating strategy, we have achieved a 0.14\% improvement in the average weekly completion rate compared to non-incremental strategies. This result further demonstrates the effectiveness of the ICE-PKD framework.

\section{Conclusion}
In this paper, we introduce the ICE-PKD framework to address the challenges of continual causal effect estimation in marketing, where covariate shift and concept drift frequently occur. Our approach extends the DR-CFR framework to accommodate multiple treatments and presents a novel incremental method for uplift modeling. 

The experimental results demonstrate that ICE-PKD consistently outperforms alternative strategies across both synthetic and online datasets. In the presence of covariate shift and concept drift, ICE-PKD can effectively ensure stable and superior performance over time. Additionally, ablation experiments highlight the crucial roles of the integrated components in enhancing the model's performance and robustness.

Overall, ICE-PKD represents a significant advancement in continual causal effect estimation, providing a more effective and robust approach for dynamic marketing environments. Several promising avenues for future research include exploring uplift models better suited for estimating continuous causal effects and investigating advanced incremental learning methods, such as enhanced knowledge distillation techniques, improved replay mechanisms and other specialized structures. 

\bibliographystyle{unsrtnat}
\bibliography{references}  






\end{document}